\documentclass{article}

\usepackage{PRIMEarxiv}

\usepackage[utf8]{inputenc} 
\usepackage[T1]{fontenc}    
\usepackage{hyperref}       
\usepackage{url}            
\usepackage{booktabs}       
\usepackage{amsfonts}       
\usepackage{nicefrac}       
\usepackage{microtype}      
\usepackage{lipsum}
\usepackage{fancyhdr}       
\usepackage{graphicx}       
\graphicspath{{media/}}     
\usepackage{amsmath}
\usepackage{algorithm2e}
\newcommand{\pluseq}{\mathrel{+}=}
\title{RRMSE Voting Regressor: A weighting function based improvement to ensemble regression
}

\author{
  Shikun Chen \\
  Mathematics for Engineers, Institute for Technologies of Metals \\
  University of Duisburg-Essen \\
  Friedrich-Ebert-Str. 12, 47119 Duisburg, Germany\\
  \texttt{shikun.chen@uni-due.de} \\
   \AND
  Nguyen Manh Luc \\
   \texttt{lucnm@fsoft.com.vn} \\
}

\begin{document}
\maketitle

\begin{abstract}
This paper describes the RRMSE (Relative Root Mean Square Error) based weights to weight the occurrences of predictive values before averaging for the ensemble voting regression. The core idea behind ensemble regression is to combine several base regression models in order to improve the prediction performance in learning problems with a numeric continuous target variable. The default weights setting for the ensemble voting regression is uniform weights, and without domain knowledge of learning task, assigning weights for predictions are impossible, which makes it very difficult to improve the predictions. This work attempts to improve the prediction of voting regression by implementing the RRMSE based weighting function. Experiments show that RRMSE voting regressor produces significantly better predictions than other state-of-the-art ensemble regression algorithms on six popular regression learning datasets. 
\end{abstract}

\keywords{RRMSE \and Supervised Machine learning\and Ensemble Regression \and Weighting Function}

\section{Introduction}
Ensemble learning refers to a machine learning paradigm where multiple models (often called "base learner") are trained to solve the same problem and combined to gain better results, either in classification or regression problems. The combination can be implemented by aggregating the output from each model with two objectives: reducing the model error and maintaining its generalization. Compared to a single model, the ensemble model brings more robustness and accuracy to final predictions \cite{garcia2005cooperative}.\par
More formally, an ensemble \( \mathcal{F} \) is composed of a set of predictors of a function $f$ denoted as $\hat{f_{i}}$.
\begin{equation}
\label{eq:1}
 \mathcal{F}  =  \{\hat{f_{i}}, i=1,\ldots, k\}
\end{equation}
The resulting ensemble predictor is denoted as $\hat{f_{f}}$. In general, the ensemble process can be divided into three steps~\cite{roli2001methods}. It starts with \textit{ensemble generation}, which consists of selecting base learner to be aggregated. If the base models are chosen using the same induction algorithm then the ensemble is called homogeneous, otherwise it is called heterogeneous. The second step is \textit{ensemble pruning}, the ensemble model is pruned by eliminating some of the models generated earlier. Finally, in the \textit{ensemble integration} step, a strategy to combine the base models is conducted. This strategy is then used to obtain the final prediction of the ensemble for new datasets, based on the predictions of the base learners~\cite{mendes2012ensemble}. More specifically, for regression problems, ensemble integration is performed using a linear combination of the predictions. 
\begin{equation}
\label{eq:2}
 \hat{f_{f}}(x)  =  \sum_{i=1}^{k}[h_{i}(x) \ast \hat{f_{i}}(x)]
\end{equation}
where $h_{i}(x)$ are the weighting functions. The integration approaches can be divided into constant and non-constant weighting functions~\cite{merz1998classification}. In the first case, the $h_{i}(x)$ are constants, on the other hand, the weights vary according to the feature values $x$. \par
This study introduces the \textit{RRMSE Voting Regressor}, using RRMSE to determine the weighting function for heterogeneous ensemble generation. The performance of this new method is compared to other relevant ensemble regression algorithms.

\section{Ensemble Learning for Regression}
\label{sec:ensemble}
There are basically two groups ensemble methods that are usually distinguished, the first one is \textit{averaging methods}, the basic principle is to build several learners independently and then to average their predictions. By contrast, in \textit{boosting methods}, base learners are built sequentially and one tries to reduce the bias of the combined learner. Since this work focuses on the weighting function for heterogeneous ensemble models, only the common averaging methods are discussed below. 

\subsection{Bagging Regression}
\begin{figure}
\centering
\includegraphics[width=\textwidth]{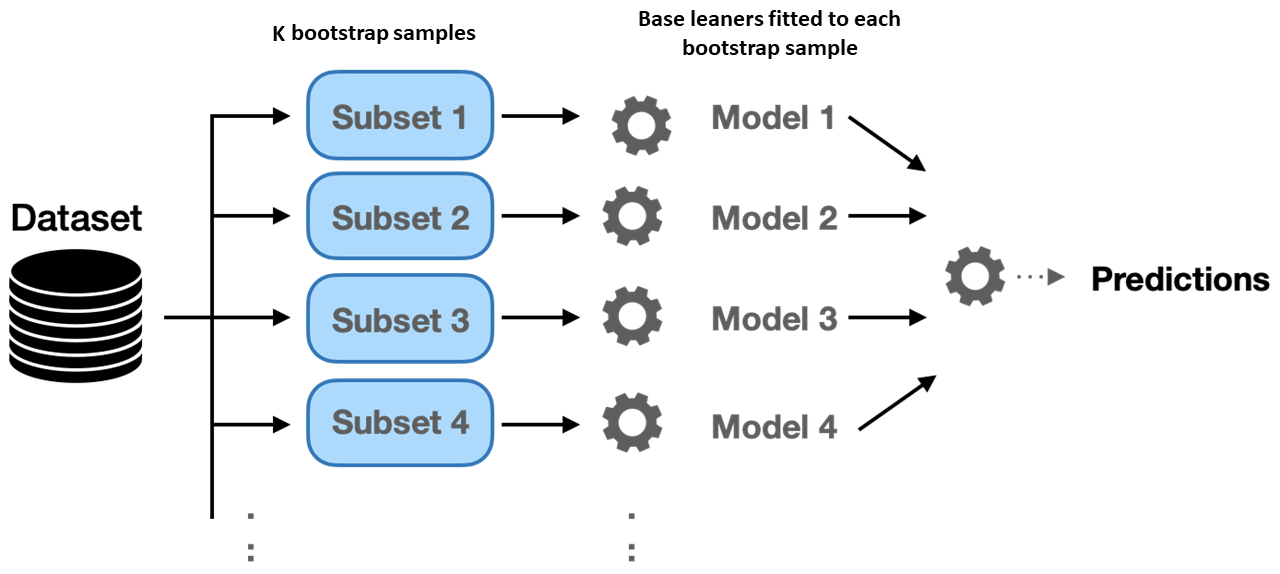}
\caption{Bagging consists in fitting several base models on different bootstrap samples and build an ensemble model that “average” the results of these base learners.}
\label{fig:1}
\end{figure}

Bagging~\cite{breiman1996bagging} is often considered as homogeneous base learners. Each model learns the error produced by the previous model using a slightly different subset of the training dataset. By introducing randomization into its construction procedure and then making an ensemble out of it, bagging is able to reduce the variance of a base learner~\cite{parmanto1996reducing}. Bagging is based on a bootstrapping sampling techniques, which creates multiple sets of the original training dataset with replacement. Each subset is of the same size and can be used to train models in parallel.  \hyperref[fig:1]{Figure 1} illustrates the workflow of the bagging algorithm. In this work, a heterogeneous based bagging algorithm was developed, so that it can be used for comparison with the RRMSE voting regressor. 

\subsection{Voting Regression}
\begin{figure}
\centering
\includegraphics[width=\textwidth]{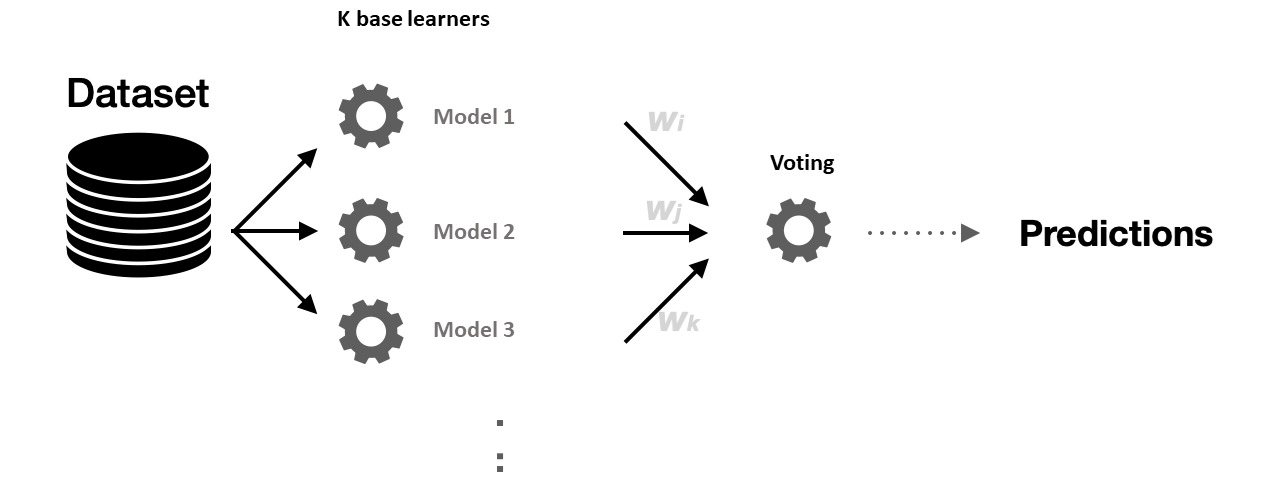}
\caption{Voting consists in fitting several base models on whole dataset and build an ensemble model that “average” the results of these base learners.}
\label{fig:2}
\end{figure}

Voting regression is another widely used ensemble method. It combines conceptually different regression models and return the average predicted values. Unlike bagging, each base model of voting regression fits on the whole dataset. The way these predictions are combined plays a key role in the final predictions. \hyperref[fig:2]{Figure 2} shows the structure of voting regression, note that every base leaner has its own weight.

\subsection{Related Work}
This section describes the related work in terms of the weighting function, i.e., ensemble integration. Equation \ref{eq:2} shows that the ensemble integration is done using a linear combination of the predictions. To be more precise, in the case of constant weighting functions, $h_i(x)$ will be replaced by the coefficient $\alpha_i$.

\subsubsection{Constant Weighting Functions}
Constant weight integration functions always use the same set of coefficients, regardless of the input to the prediction. ~\cite{perrone1993textordfemininewhen} propose the Basic Ensemble Method (BEM), which uses as an estimator for the target function.
\begin{equation}
\label{eq:3}
 \hat{f}_{BEM}(x)  =  f(x) - \frac{1}{k}\sum_{i=1}^{k}m_i(x)
\end{equation}
where
\begin{equation}
\label{eq:4}
 m_i(x) = f(x) - \hat{f}_i(x)
\end{equation}

\noindent BEM assumes that the $m_i(x)$ are mutually independent with zero mean. To solve this issue, ~\cite{perrone1993textordfemininewhen} modify the estimator to GBM - Generalized Ensemble Method. 
\begin{equation}
\label{eq:5}
 \hat{f}_{GEM}(x)  =  \sum_{i=1}^{k}[\alpha_i \ast \hat{f}_i(x)] = f(x) + \sum_{i=1}^{k}[\alpha_i \ast m_i(x)] 
\end{equation}
where
\begin{equation}
\label{eq:6}
\begin{split}
 \sum_{j=1}^{k}\alpha_i = 1, \\
 \alpha_i = \frac{\sum_{j=1}^{k}C_{ij}^{-1}}{\sum_{l=1}^{k}\sum_{j=1}^{k}C_{ij}^{-1}}, \\
 C_{ij} = E[m_i(x) \ast m_j(x)]
\end{split}
\end{equation}
\noindent The disadvantage of this approach is the multi-collinearity problem, which causes the confidence intervals for the $\alpha_i$ coefficients to be wide, i.e., the estimators of the coefficients will have high variance ~\cite{merz1998classification}, since it is necessary to calculate the inverse matrix $C^{-1}$. To avoid multi-collinearity problem, ~\cite{caruana2004ensemble} embed the ensemble integration phase in the ensemble selection phase. By selecting models from the pool to be included in the ensemble as a substitute and using the simple average as the weighting function, the $\alpha_i$ coefficients are implicitly calculated as the number of times each model is selected relative to the total number of models in the ensemble. ~\cite{merz1999principal} take the advantage of principal component regression (PCR) to prevent the multi-collinearity problem. The PCR method determines the principal components (PCs) and then selects the number of PCs to use. The search of the number of PCs to apply is then based on the ordered PCs, and it is a function of the variation. 

\subsubsection{Non-Constant Weighting Functions}
The difference between a constant and a non-constant weighting function is that the non-constant function uses different coefficients for prediction depending on the input. The predefined coefficients based on prior domain knowledge or areas of expertise for each model are called static. On the other hand, in the dynamic approach, the $h_i(x)$ weights from equation \ref{eq:2} are determined based on the performance of the base learners. Predictors are usually selected by evaluating their performance against similar data from the training set for a selected performance measure. ~\cite{woods1997combination} and ~\cite{puuronen1999dynamic} use k-nearest neighbors with the Euclidean distance and weighted k-nearest neighbors to choose predictors. Dynamic weighting (DW) ~\cite{Rooney2004DynamicIO} assigns a weight to each base learner according to its localized performance in the group of k-nearest neighbors and the final prediction is based on the weighted average of the predictions of the associated models. In this work, a dynamic weighting voting regressor (DWR) was implemented to compare its performance with the RRMSE voting regressor.

\section{RRMSE Voting Regressor}
\label{sec:rrmse}
In this section, the definition of RRMSE (Relative Root Mean Squared Error) will be provided firstly, and then the algorithm of RRMSE voting regressor will be presented. \par

RRMSE is derived from RMSE (Root Mean Squared Error), which is defined as follow:
\begin{equation}
\label{eq:7}
    \text{RMSE}(f, \hat{f}) = \sqrt{\frac{1}{n} \sum_{i=0}^{n-1} (f_i - \hat{f}_i)^2}
\end{equation}
where $\hat{f}_i$ is the predicted value of the $i$-th sample, and $f_i$ is the corresponding actual value, $n$ is the number of samples, when divide the difference between the predicted value and actual value $f_i - \hat{f}_i$ by $\hat{f}_i$, then error will be considered relatively, which means that each residual is scaled against actual value.
\begin{equation}
\label{eq:8}
\text{RRMSE}(f, \hat{f}) = \sqrt{\frac{\frac{1}{n}\sum_{i=0}^{n-1}(f_i-\hat{f}_i)^2}{\sum_{i=0}^{n-1}\hat{f}_i^2}}
\end{equation}
A good ensemble is the one with accurate predictors and making errors in different parts of the input space. For the regression problem it is possible to decompose the generalization error in different components, using RRMSE as dynamic weighting function can guide the process to optimize the ensemble integration. \par
A high-level description of the proposed approach is presented in Algorithm \ref{alg:1}. In the first phase, a constant number is calculated to avoid zero division. Subsequently, the RRMSE of each based learner are evaluated base on equation \ref{eq:8}. In the third phase, the weights are assigned inversely proportional to their total sum. Finally, the final predictions on testing dataset combine the individual predictions of chosen base learners utilizing the proposed weighting function. 
\RestyleAlgo{ruled}
\begin{algorithm}
\caption{RRMSE Voting Regressor}\label{alg:1}
\SetKwInOut{Input}{Input}
\SetKwInOut{Output}{Output}
\SetKwInput{kwInit}{Init}
\Input{K - Set of base learners. \\
        X - Set of training features of size $n\times m$ \\
        T - Set of testing data \\
        $y$ - Set of training labels of size $n$ \\
        $y'$ - Set of predicted value on each base learner \\
    }
\Output{$\hat{y}$ - Set of predicted values}
\BlankLine
\textbf{Phase 1: calculate a constant to prevent zero division.} \\
\kwInit {$\text{avg} \gets mean(y)$, \quad $\text{sum} \gets 0$}
\For{$i$ \KwTo $n$} 
{$\text{sum} \pluseq |(y[i] - \text{avg})|$}
$\text{constant} \gets  \text{sum} \div n $ \\
\BlankLine
\textbf{Phase 2: calculate set of RRMSE.} \\
\kwInit{$\text{result} \gets 0$}
\ForEach{$i \in n$}
{
    $\text{diff} \gets y[i] - y'[i] $ \;
    $\text{denominator} \gets y[i] + \text{constant}$ \;
    $\text{result} \pluseq (\text{diff} \div \text{denominator})^2$
}
$\text{rrmse} \gets \sqrt{\text{result} \div n}$
\BlankLine
\textbf{Phase 3: assigning RRMSE weights to each learner} \\
\kwInit{$\text{errors} \gets \text{empty array []} \quad $W$ \gets \text{empty array []}$}
\ForEach{$k \in K$}
{
    $\hat{y}_k \gets \text{predict label using X}$ \;
    calculate the constant and rrmse for each $k$ \;
    append each rrmse to errors \;
}
$\text{errorList} \gets \text{inverse the element in errors}$; \\
$\text{sumError} \gets \text{sum of all element in errorList}$; \\
\ForEach{$j \in \text{errorList}$}{
    $\text{$W$.insert}($j$ \div \text{sumError})$
}

\BlankLine
\textbf{Phase 4: RRMSE weighted voting prediction} \\
\ForEach{$x \in T$}{
\For{$i=1$ \KwTo K}{
    Apply regressor $K_i$ on $x$.
    }
Predict the label $\hat{y}$ of $x$ using \\
$\hat{y}  =  \sum_{i=1}^{k}[W_{i}(x) \ast K_{i}(x)]$
    
}
\end{algorithm}

\section{Experiments}
\label{sec:experiments}
To access the effectiveness of RRMSE voting regressor, experiments were performed using six well known regression datasets, list in Table \ref{tab:table1}. \textit{Instances} and \textit{Features} indicate the total number of data points and the number of predictor variables. The Abalone dataset ~\cite{Dua:2019} consists of 8 physical measurements, and the task is to predict the age of Abalone shells. The Car dataset ~\cite{Dua:2019} contains 7 features and the predicted attribute is "mpg", which means the fuel consumption per gallon. The Diamond dataset contains the prices and other 9 attributes of 53940 diamonds. Airfoil dataset ~\cite{Dua:2019} obtained from a series of aerodynamic and acoustic tests of two and three-dimensional airfoil blade section conducted in an anechoic wind tunnel. The task of Airfoil is to predict the scaled sound pressure level by using 5 input features. Dataset Smart Grid Stability ~\cite{Dua:2019} includes the data of local stability analysis of the 4-node star system implementing Decentral Smart Grid Control concept. The Elongation dataset has 17 chemical components as input attributes, the task is to predict the elongation of steel. \par
To evaluate the performance of the RRMSE voting regressor, three state-of-the-art ensemble regression algorithms were also introduced. The first one is voting regressor with uniform weights (VRU), is implemented in the Scikit-Learn library ~\cite{scikit-learn}. The other two are heterogeneous based bagging regressor with uniform weights (BR) and dynamic weighting regressor (DWR) \cite{Rooney2004DynamicIO}.
RRMSE, BR and DWR were implemented by the author in Python. \par
In addition to the RMSE given in equation \ref{eq:7}, to measure the model performance for regression task, the following metrics are also presented. Mean absolute error (MAE) is a loss metric corresponding to the expected value of the absolute error loss. If $\hat{f}_i$ is the predicted value of the $i$-th sample, and $f_i$ is the corresponding true value, $n$ is the number of sample, the MAE estimated over $n$ is defined as:
\begin{equation}
\label{eq:9}
    \text{MAE}(f, \hat{f}) =\frac{1}{n} \sum_{i=0}^{n-1} |f_i - \hat{f}_i|
\end{equation}
Mean squared error (MSE): is a loss metric corresponding to the expected value of the squared error. MSE estimated over $n$ is defined as:
\begin{equation}
\label{eq:10}
    \text{MSE}(f, \hat{f}) =\frac{1}{n} \sum_{i=0}^{n-1} (f_i - \hat{f}_i)^2
\end{equation}
Finally, $R^2$ score represents the proportion of variance of $f$ that has been explained by the independent variables in the model. It provides an indication of fitting goodness and therefore a measure of how well unseen samples are likely to be predicted by the model:

 \begin{equation}
 \label{eq:11}
        R^2(f, \hat{f}) = 1 - \dfrac{\sum\limits_{i=0}^{n-1} (f_i - \hat{f}_i)^2}{\sum\limits_{i=0}^{n-1} (f_i - \bar{f})^2}
\end{equation}

\begin{table}
\begin{center}
\caption {The datasets used in the experiments described in this paper and their properties} \label{tab:table1} 
\smallskip
\begin{tabular}{llllll}
\multicolumn{1}{c}{Dataset} & \multicolumn{1}{c}{Domain}   & \multicolumn{1}{c}{Instances} & \multicolumn{1}{c}{Features}  \\ \hline
\multicolumn{1}{c}{Abalone} & \multicolumn{1}{c}{Zoology} & \multicolumn{1}{c}{4177}     & \multicolumn{1}{c}{8} \\ 
\multicolumn{1}{c}{Car} & \multicolumn{1}{c}{Automobile} & \multicolumn{1}{c}{398}     & \multicolumn{1}{c}{7}    \\ 
\multicolumn{1}{c}{Diamond} & \multicolumn{1}{c}{Consumption} & \multicolumn{1}{c}{53940}     & \multicolumn{1}{c}{9}     \\ 
\multicolumn{1}{c}{Airfoil} & \multicolumn{1}{c}{Aerodynamic} & \multicolumn{1}{c}{1503}     & \multicolumn{1}{c}{5}     \\ 
\multicolumn{1}{c}{Smart Grid Stability} & \multicolumn{1}{c}{Electricity} & \multicolumn{1}{c}{60000}     & \multicolumn{1}{c}{12}     \\ 
\multicolumn{1}{c}{Elongation} & \multicolumn{1}{c}{Metallurgy} & \multicolumn{1}{c}{385}     & \multicolumn{1}{c}{17}     \\ 
\end{tabular}
\end{center}
\end{table}

For each ensemble algorithm, the same set of base learners are chosen, they are Linear Regression (LR), K-Nearest Neighbors Regression (KNN), Stochastic Gradient Descent Regression (SGD) and Random Forest Regression (RF). These four regression models are implemented in the Scikit-Learn library ~\cite{scikit-learn}. Meanwhile, $80\%$ dataset are split for training, and $20\%$ are used as testing dataset. 

\section{Results}
\label{sec:results}
To show the performance of the RRMSE voting regressor and compare it with other methods, the hyperparameters for all base learners are set to default values. Table \ref{tab:table2} shows the 4 regression metrics and their performances across different testing datasets. The integer in braces after number indicates the ranking for that specific dataset over the different algorithms, the number after algorithm names indicate the average rank of that algorithm with respect to all metrics (where a lower value indicates a better rank). \par
From Table \ref{tab:table2} it can be seen that the RRMSE voting regressor performs better than the other 3 algorithms over 5 datasets, except for Abalone dataset (average rank 2). To assess the degree of difference between the performances of all 4 algorithms, non-parametric statistical significance tests were performed. ~\cite{garcia2010advanced} propose \textit{Friedman aligned rank test} to compare different algorithms among a set of different datasets. The significance test indicates a difference among the algorithms with a significance level of $\alpha = 0.05$. Next, a pairwise post-hoc \textit{Friedman aligned rank test} was performed to explore which algorithm had differed significantly. \par
The comparisons between algorithms and the p-values from the pairwise post-hoc Friedman aligned rank tests are presented in Table \ref{tab:table3} which is partitioned in two parts. A cell in the upper-diagonal of the table indicates the win/lose/tie counts of the algorithm in the corresponding row with respect to the algorithm in the corresponding column. This is included for a direct comparison between algorithm pairs. For instance. RRMSE voting regressor has better results than VRU in 5 of the datasets and worse in 1 dataset. The lower diagonal part of the Table \ref{tab:table3} shows the p-values of the post-hoc Friedman aligned rank tests. The asterisk (*) indicates if the comparison was found to be significantly different. The different levels of significance ($\alpha$) are defined by the number of asterisks, where * means $\alpha=0.10$, ** means $\alpha=0.05$ and *** indicates $\alpha=0.01$.\par

It can be seen that RRMSE voting regressor performed better over all datasets than DWR with all three significance levels, and better than BR with a significance level of 0.05. The null hypothesis could be rejected with all three significance level for VRU. \par

\begin{table}
\begin{center}
\caption {Four regression metrics illustrating performance of the different algorithms compared in this study across the different testing datasets used. Number in brackets indicates relative ranking of the algorithm.} \label{tab:table2} 
\smallskip
\begin{tabular}{llllll}

\multicolumn{1}{c}{Dataset} & \multicolumn{1}{c}{Methods}   & \multicolumn{1}{c}{MAE} & \multicolumn{1}{c}{MSE} & \multicolumn{1}{c}{RMSE}  & \multicolumn{1}{c}{$R^2$}  \\ \hline

\multicolumn{1}{c}{Abalone} & \multicolumn{1}{c}{RRMSE (2)} & \multicolumn{1}{c}{1.5102 (2)}     & \multicolumn{1}{c}{4.7140 (2)} & \multicolumn{1}{c}{2.1712 (2)} & \multicolumn{1}{c}{0.5645 (2)} \\ 
\multicolumn{1}{c}{} & \multicolumn{1}{c}{BR (3)} & \multicolumn{1}{c}{1.5290 (3)}     & \multicolumn{1}{c}{4.7995 (3)}  & \multicolumn{1}{c}{2.1908 (3)}   & \multicolumn{1}{c}{0.5566 (3)} \\ 
\multicolumn{1}{c}{} & \multicolumn{1}{c}{VRU (1)} & \multicolumn{1}{c}{1.4949 (1)}     & \multicolumn{1}{c}{4.6174 (1)}   & \multicolumn{1}{c}{2.1488 (1)}   & \multicolumn{1}{c}{0.5735 (1)} \\ 
\multicolumn{1}{c}{} & \multicolumn{1}{c}{DWR (4)} & \multicolumn{1}{c}{1.5853 (4)}     & \multicolumn{1}{c}{4.8962 (4)}   & \multicolumn{1}{c}{2.2127 (4)}   & \multicolumn{1}{c}{0.5477 (4)} \\ 
\hline

\multicolumn{1}{c}{Car} & \multicolumn{1}{c}{RRMSE (1)} & \multicolumn{1}{c}{1.6758 (1)}     & \multicolumn{1}{c}{4.7219 (1)} & \multicolumn{1}{c}{2.1730 (1)} & \multicolumn{1}{c}{0.9122 (1)} \\ 
\multicolumn{1}{c}{} & \multicolumn{1}{c}{BR (3)} & \multicolumn{1}{c}{1.9065 (3)}     & \multicolumn{1}{c}{6.0156 (3)}  & \multicolumn{1}{c}{2.4527 (3)}   & \multicolumn{1}{c}{0.8881 (3)} \\ 
\multicolumn{1}{c}{} & \multicolumn{1}{c}{VRU (2)} & \multicolumn{1}{c}{1.8426 (2)}     & \multicolumn{1}{c}{5.4789 (2)}   & \multicolumn{1}{c}{2.3407 (2)}   & \multicolumn{1}{c}{0.8981 (2)} \\ 
\multicolumn{1}{c}{} & \multicolumn{1}{c}{DWR (4)} & \multicolumn{1}{c}{2.4460 (4)}     & \multicolumn{1}{c}{9.6812 (4)}   & \multicolumn{1}{c}{3.1115 (4)}   & \multicolumn{1}{c}{0.8199 (4)} \\ 
\hline

\multicolumn{1}{c}{Diamond} & \multicolumn{1}{c}{RRMSE (1)} & \multicolumn{1}{c}{301.8838 (1)}     & \multicolumn{1}{c}{330903 (1)} & \multicolumn{1}{c}{575.24 (1)} & \multicolumn{1}{c}{0.9792 (1)} \\ 
\multicolumn{1}{c}{} & \multicolumn{1}{c}{BR (3)} & \multicolumn{1}{c}{449.6199 (3)}     & \multicolumn{1}{c}{615979 (3)}  & \multicolumn{1}{c}{784.65 (3)}   & \multicolumn{1}{c}{0.9613 (3)} \\ 
\multicolumn{1}{c}{} & \multicolumn{1}{c}{VRU (2)} & \multicolumn{1}{c}{437.4929 (2)}     & \multicolumn{1}{c}{571903 (2)}   & \multicolumn{1}{c}{756.24 (2)}   & \multicolumn{1}{c}{0.9640 (2)} \\ 
\multicolumn{1}{c}{} & \multicolumn{1}{c}{DWR (4)} & \multicolumn{1}{c}{718.8416 (4)}     & \multicolumn{1}{c}{1473423 (4)}   & \multicolumn{1}{c}{1213.84 (4)}   & \multicolumn{1}{c}{0.9073 (4)} \\ 
\hline

\multicolumn{1}{c}{Airfoil} & \multicolumn{1}{c}{RRMSE (1)} 
& \multicolumn{1}{c}{1.8090 (1)}     & \multicolumn{1}{c}{5.3482 (1)} & \multicolumn{1}{c}{2.3126 (1)} & \multicolumn{1}{c}{0.8932 (1)} \\ 
\multicolumn{1}{c}{} & \multicolumn{1}{c}{BR (3)} 
& \multicolumn{1}{c}{3.1896 (3)}     & \multicolumn{1}{c}{16.0696 (3)}  & \multicolumn{1}{c}{4.0074 (3)}   & \multicolumn{1}{c}{0.6794 (3)} \\ 
\multicolumn{1}{c}{} & \multicolumn{1}{c}{VRU (2)} 
& \multicolumn{1}{c}{3.0898 (2)}     & \multicolumn{1}{c}{14.9672 (2)}   & \multicolumn{1}{c}{3.8688 (2)}   & \multicolumn{1}{c}{0.7012 (2)} \\ 
\multicolumn{1}{c}{} & \multicolumn{1}{c}{DWR (4)} 
& \multicolumn{1}{c}{3.6852 (4)}     & \multicolumn{1}{c}{22.9774 (4)}   & \multicolumn{1}{c}{4.7935 (4)}   & \multicolumn{1}{c}{0.5414 (4)} \\ 
\hline
\multicolumn{1}{c}{Smart Grid Stability} & \multicolumn{1}{c}{RRMSE (1)} 
& \multicolumn{1}{c}{0.0075 (1)}     & \multicolumn{1}{c}{0.000095 (1)} & \multicolumn{1}{c}{0.0098 (1)} & \multicolumn{1}{c}{0.9287 (1)} \\ 
\multicolumn{1}{c}{} & \multicolumn{1}{c}{BR (3)} 
& \multicolumn{1}{c}{0.0116 (3)}     & \multicolumn{1}{c}{0.0002 (3)}  & \multicolumn{1}{c}{0.0146 (3)}   & \multicolumn{1}{c}{0.8414 (3)} \\ 
\multicolumn{1}{c}{} & \multicolumn{1}{c}{VRU (2)} 
& \multicolumn{1}{c}{0.0112 (2)}     & \multicolumn{1}{c}{0.0001 (2)}   & \multicolumn{1}{c}{0.0140 (2)}   & \multicolumn{1}{c}{0.8541 (2)} \\ 
\multicolumn{1}{c}{} & \multicolumn{1}{c}{DWR (4)} 
& \multicolumn{1}{c}{0.0174 (4)}     & \multicolumn{1}{c}{0.0004 (4)}   & \multicolumn{1}{c}{0.0219 (4)}   & \multicolumn{1}{c}{0.6425 (4)} \\ 
\hline
\multicolumn{1}{c}{Elongation} & \multicolumn{1}{c}{RRMSE (1)} 
& \multicolumn{1}{c}{0.8219 (1)}     & \multicolumn{1}{c}{1.1775 (1)} & \multicolumn{1}{c}{1.0851 (1)} & \multicolumn{1}{c}{0.6271 (1)} \\ 
\multicolumn{1}{c}{} & \multicolumn{1}{c}{BR (2.75)} 
& \multicolumn{1}{c}{0.8524 (2)}     & \multicolumn{1}{c}{1.3431 (3)}  & \multicolumn{1}{c}{1.1589 (3)}   & \multicolumn{1}{c}{0.5746 (3)} \\ 
\multicolumn{1}{c}{} & \multicolumn{1}{c}{VRU (2.25)} 
& \multicolumn{1}{c}{0.8531 (3)}     & \multicolumn{1}{c}{1.3174 (2)}   & \multicolumn{1}{c}{1.1478 (2)}   & \multicolumn{1}{c}{0.5828 (2)} \\ 
\multicolumn{1}{c}{} & \multicolumn{1}{c}{DWR (4)} 
& \multicolumn{1}{c}{0.9984 (4)}     & \multicolumn{1}{c}{1.6482 (4)}   & \multicolumn{1}{c}{1.2838 (4)}   & \multicolumn{1}{c}{0.4780 (4)} \\ 
\hline
\end{tabular}
\end{center}
\end{table}

In summary, the results indicate that RRMSE voting regressor was able to perform better than BR and DWR. Although it was found not be very different than VRU, based on the test, RRMSE voting regressor achieved the best average rank. 

\section{Summary and Conclusion}
\label{sec:summary}
In this paper, a new weighted voting ensemble algorithm for the regression problem, entitled RRMSE voting regressor, was proposed. The weighting function for base learners is assigned based on the inverse proportion of their sum of RRMSE. Statistical significance tests show that the proposed method was able to perform significantly better than BR and DWR, it seems to also be better than VRU, but only marginally. \par
The results in Section \ref{sec:results} show that the RRMSE voting regressor did improve the performance of ensemble regression and therefore it would now be interesting to further explore RRMSE voting regressor with the focus of improving the predictions. \par
There are many alternatives for base learner, and the setting of hyperparameters for the individual base learner could also be enormous. The results motivate the author to extend the method in the direction of better base learner selection, using Random Search or Bayesian Optimization to improve the prediction quality of base learners.

\begin{table}
\begin{center}
\caption {Significance test on metric MAE, MSE, RMSE and $R^2$\\
Upper diagonal: win/lose/tie. Lower diagonal: post-hoc Friedman Aligned Rank\\
Test p-values: Significance levels: $\ast$ : $\alpha = 0.1$, $\ast \ast$ : $\alpha = 0.05$, $\ast \ast \ast$ : $\alpha=0.01$} \label{tab:table3} 
\smallskip
\begin{tabular}{cccccc}
\multicolumn{1}{c}{MAE} & \multicolumn{1}{c}{RRMSE}   & \multicolumn{1}{c}{BR} & \multicolumn{1}{c}{VRU}  & \multicolumn{1}{c}{DWR} \\ \hline
\multicolumn{1}{c}{RRMSE} & \multicolumn{1}{c}{} & \multicolumn{1}{c}{6/0/0}     & \multicolumn{1}{c}{5/0/1}
& \multicolumn{1}{c}{6/0/0} \\ 
\multicolumn{1}{c}{BR} & \multicolumn{1}{c}{0.2529} & \multicolumn{1}{c}{}     & \multicolumn{1}{c}{1/5/0}  & \multicolumn{1}{c}{6/0/0}  \\ 
\multicolumn{1}{c}{VRU} & \multicolumn{1}{c}{0.4142} & \multicolumn{1}{c}{0.7439}     & \multicolumn{1}{c}{}   & \multicolumn{1}{c}{6/0/0}   \\ 
\multicolumn{1}{c}{DWR} & \multicolumn{1}{c}{*** 0.0011} & \multicolumn{1}{c}{** 0.0337}     & \multicolumn{1}{c}{** 0.0143}     \\ 
\hline
\multicolumn{1}{c}{MSE} & \multicolumn{1}{c}{RRMSE}   & \multicolumn{1}{c}{BR} & \multicolumn{1}{c}{VRU}  & \multicolumn{1}{c}{DWR} \\ \hline
\multicolumn{1}{c}{RRMSE} & \multicolumn{1}{c}{} & \multicolumn{1}{c}{6/0/0}     & \multicolumn{1}{c}{5/0/1}
& \multicolumn{1}{c}{6/0/0} \\ 
\multicolumn{1}{c}{BR} & \multicolumn{1}{c}{0.1779} & \multicolumn{1}{c}{}     & \multicolumn{1}{c}{0/6/0}  & \multicolumn{1}{c}{6/0/0}  \\ 
\multicolumn{1}{c}{VRU} & \multicolumn{1}{c}{0.4379} & \multicolumn{1}{c}{0.5676}     & \multicolumn{1}{c}{}   & \multicolumn{1}{c}{6/0/0}   \\ 
\multicolumn{1}{c}{DWR} & \multicolumn{1}{c}{*** 0.0010} & \multicolumn{1}{c}{** 0.0550}     & \multicolumn{1}{c}{** 0.0127}     \\ 
\hline
\multicolumn{1}{c}{RMSE} & \multicolumn{1}{c}{RRMSE}   & \multicolumn{1}{c}{BR} & \multicolumn{1}{c}{VRU}  & \multicolumn{1}{c}{DWR} \\ \hline
\multicolumn{1}{c}{RRMSE} & \multicolumn{1}{c}{} & \multicolumn{1}{c}{6/0/0}     & \multicolumn{1}{c}{5/0/1}
& \multicolumn{1}{c}{6/0/0} \\ 
\multicolumn{1}{c}{BR} & \multicolumn{1}{c}{0.1779} & \multicolumn{1}{c}{}     & \multicolumn{1}{c}{0/6/0}  & \multicolumn{1}{c}{6/0/0}  \\ 
\multicolumn{1}{c}{VRU} & \multicolumn{1}{c}{0.4379} & \multicolumn{1}{c}{0.5676}     & \multicolumn{1}{c}{}   & \multicolumn{1}{c}{6/0/0}   \\ 
\multicolumn{1}{c}{DWR} & \multicolumn{1}{c}{*** 0.0010} & \multicolumn{1}{c}{** 0.0550}     & \multicolumn{1}{c}{** 0.0127}     \\ 
\hline
\multicolumn{1}{c}{$R^2$} & \multicolumn{1}{c}{RRMSE}   & \multicolumn{1}{c}{BR} & \multicolumn{1}{c}{VRU}  & \multicolumn{1}{c}{DWR} \\ \hline
\multicolumn{1}{c}{RRMSE} & \multicolumn{1}{c}{} & \multicolumn{1}{c}{6/0/0}     & \multicolumn{1}{c}{5/0/1}
& \multicolumn{1}{c}{6/0/0} \\ 
\multicolumn{1}{c}{BR} & \multicolumn{1}{c}{** 0.04122} & \multicolumn{1}{c}{}     & \multicolumn{1}{c}{0/6/0}  & \multicolumn{1}{c}{6/0/0}  \\ 
\multicolumn{1}{c}{VRU} & \multicolumn{1}{c}{0.2884} & \multicolumn{1}{c}{0.3271}     & \multicolumn{1}{c}{}   & \multicolumn{1}{c}{6/0/0}   \\ 
\multicolumn{1}{c}{DWR} & \multicolumn{1}{c}{*** 0.00008} & \multicolumn{1}{c}{** 0.0603}     & \multicolumn{1}{c}{*** 0.00426}     \\ 
\end{tabular}
\end{center}
\end{table}

\section*{Acknowledgments}
This research has received no external funding.
\bibliographystyle{unsrt}  
\bibliography{references}

\end{document}